\documentclass[twoside,11pt]{article}
\usepackage{jair, apalike, rawfonts}
\usepackage{graphicx}
\usepackage{enumitem}
\usepackage[export]{adjustbox}[2011/08/13]
\usepackage{caption}
\usepackage[utf8x]{inputenc}
\usepackage{pgf}
\usepackage{hanging}
\usepackage{verbatim}

\usepackage[T1]{fontenc}    % use 8-bit T1 fonts
\usepackage{url}            % simple URL typesetting
\usepackage{amsfonts}       % blackboard math symbols
\usepackage{nicefrac}       % compact symbols for 1/2, etc.
\usepackage{microtype}      % microtypography

\usepackage{makecell} % To keep spacing of text in tables
\setcellgapes{4pt} % parameter for the spacing

\ShortHeadings{Unity: A General Platform for Intelligent Agents}
{Juliani, Berges, Teng, Cohen, Harper, Elion, Goy, Gao, Henry, Mattar, \& Lange}
\firstpageno{1}

\begin{document}

\title{Unity: A General Platform for Intelligent Agents}

\author{\name Arthur Juliani \email arthurj@unity3d.com \\
       \name Vincent-Pierre Berges \email vincentpierre@unity3d.com \\
       \name Ervin Teng \email ervin@unity3d.com \\
       \name Andrew Cohen \email andrew.cohen@unity3d.com \\
       \name Jonathan Harper \email jharper@unity3d.com \\
       \name Chris Elion \email chris.elion@unity3d.com \\
       \name Chris Goy \email christopherg@unity3d.com \\
       \name Yuan Gao \email vincentg@unity3d.com \\
       \name Hunter Henry \email brandonh@unity3d.com \\
       \name Marwan Mattar \email marwan@unity3d.com \\
       \name Danny Lange \email dlange@unity3d.com \\
       \addr Unity Technologies\\
       San Francisco, CA  94103 USA
       }

\maketitle

\begin{abstract}
Recent advances in artificial intelligence have been driven by the presence of increasingly realistic and complex simulated environments. However, many of the existing environments provide either unrealistic visuals, inaccurate physics, low task complexity, restricted agent perspective, or a limited capacity for interaction among artificial agents. Furthermore, many platforms lack the ability to flexibly configure the simulation, making the simulated environment a black-box from the perspective of the learning system. In this work, we propose a novel taxonomy of existing simulation platforms and discuss the highest level class of {\it general platforms} which enable the development of learning environments that are rich in visual, physical, task, and social complexity. We argue that modern game engines are uniquely suited to act as general platforms and as a case study examine the Unity engine and open source Unity ML-Agents Toolkit\footnote{\url{https://github.com/Unity-Technologies/ml-agents}}. We then survey the research enabled by Unity and the Unity ML-Agents Toolkit, discussing the kinds of research a flexible, interactive and easily configurable general platform can facilitate.
\end{abstract}

\section{Introduction}

In recent years, there have been significant advances in the state of deep reinforcement learning research and algorithm design \cite{mnih2015human,schulman2017proximal,silver2017mastering,espeholt2018impala}. Essential to this rapid development has been the presence of challenging and scalable simulation platforms such as the Arcade Learning Environment \cite{bellemare2013arcade}, VizDoom \cite{kempka2016vizdoom}, MuJoCo \cite{todorov2012mujoco}, and many others \cite{beattie2016deepmind,johnson2016malmo,coumans2016pybullet}. The Arcade Learning Environment (ALE), for example, was essential for providing a means of benchmarking the control-from-pixels approach of the Deep Q-Network \cite{mnih2013playing}. Similarly, other environments and platforms have helped motivate research into more efficient and powerful algorithms \cite{oh2016control,andrychowicz2017hindsight}. The simulation environment is the fundamental way in which the reinforcement learning community tests its ideas and algorithms. Thus, the {\it quality of environments} is of critical importance. Surprisingly, the general discussion around this integral component is underdeveloped compared to its algorithmic counterpart.

Many of the current research platforms are based on popular video games or game engines such as Atari 2600, Quake III, Doom, and Minecraft. This is part of a much longer-term trend in which games have served as a platform for artificial intelligence (AI) research. This trend can be traced back to the earliest work in AI around playing games such as chess and checkers \cite{shannon1950xxii,samuel1959some}, or later work applying reinforcement learning to the game of Backgammon \cite{tesauro1995temporal}. The necessary search, decision-making and planning which make video games engaging challenges for humans are also the same challenges which interest AI researchers \cite{laird2001human}. This insight has motivated a wide range of research into the intersection of video games and AI from the diverse perspectives of game playing, player modeling, and content generation \cite{yannakakis2018artificial}. 

As deep reinforcement learning algorithms becomes more sophisticated, existing environments and the benchmarks based on them become less informative. For example, most environments in the ALE have been solved to above human-level performance, making the continued use of the benchmark less valuable \cite{machado2017revisiting,puigdomenech2020agent57}. A complementary point created by this state of algorithmic progress is that there exists a virtuous circle in which the development of novel environments drives the development of novel algorithms. We can expect the research community to continue to provide high-quality algorithms. However, it is unclear from where researchers should expect high-quality environments, since the creation of such environments is often time-intensive and requires specialized domain knowledge. This continual need for novel environments necessitates an easy-to-use, flexible and universal platform for unrestricted environment creation.

Simulated environments are constrained by the limitations of the simulators themselves. Simulators are not equal in their ability to provide meaningful challenges to learning systems. Furthermore, it is sometimes not obvious which properties of an environment make it a worthwhile benchmark for research. The complexity of the physical world is a primary candidate for  challenging the current as well as to-be-developed algorithms. It is in the physical world where mammalian and, more specifically, human intelligence developed, and it is this kind of intelligence which researchers are often interested in replicating \cite{lake2017building}. 

Modern game engines are powerful tools for the simulation of visually realistic worlds with sophisticated physics and complex interactions between agents with varying capacities. Additionally, engines designed for game development provide user interfaces which are specifically engineered to be intuitive, easy to use, interactive, and available across many platforms. Thus, in this paper we argue that game engines are perfectly poised to yield the necessary challenges for the foreseeable future of AI research. For the community, this would provide the ability to test algorithms in domains with as much depth and diversity as today's video games.

The {\bf contributions} of this work are:
\begin{itemize}
    \item A novel taxonomy of existing platforms used for research which classifies platforms in terms of their potential for complexity along the dimensions of  sensory, physical, task-logic and social.
    \item A detailed analysis of the Unity game engine and the Unity ML-Agents Toolkit as an instance of a {\it general platform}, the highest level of the proposed taxonomy.
    \item A survey of current research conducted using Unity and critical areas in which progress is hindered by the current platforms but can be facilitated by a general platform such as Unity.
\end{itemize}

This paper is structured as follows: We begin with an analysis of the properties of a simulator important for the development of learning algorithms. Then, we propose a taxonomy of simulation platforms which we use to organize popular reinforcement learning (RL) benchmarks and further point out their limitations at fully realizing all desirable properties of a simulator. We then present the Unity engine and Unity ML-Agents Toolkit a general platform and discuss the extent to which it possesses the desired characteristics for enabling research. We next outline the architecture, functionality and tools provided by the open source Unity ML-Agents Toolkit which enable the deployment of learning algorithms within Unity environments and provide a set of benchmark results on example learning environments. We conclude by proposing future avenues of research we believe will be enabled by using a flexible game engine versus standard black box environments.

\section{Anatomy of Environments and Simulators}\label{simulator_properties}

In this section, we detail some of the characteristics of environments and simulators we believe are needed to advance the state of the field in AI research.
We use the term {\it environment} to refer to the space in which an artificial agent acts and {\it simulator} to refer to the platform which computes the environment.

\subsection{Environment Properties}

As algorithms are able to solve increasingly difficult tasks, the complexity of the environments themselves must increase in order to continue to provide meaningful challenges. The specific axes of environmental complexity we believe are essential are sensory, physical, task logic, and social.  In this subsection, we outline the role each of these play in the state of the art in AI. 

{\bf Sensory Complexity} -  
The recent advances in deep learning have largely been driven by the ability of neural networks to process large amounts of visual, auditory, and text-based data \cite{lecun2015deep}. ImageNet, a large database of natural images with associated labels, was essential in enabling models such as ResNet \cite{he2016deep}, and Inception \cite{szegedy2016rethinking} to be trained to near human-level object-recognition performance \cite{russakovsky2015imagenet}. While ImageNet was mainly used for static image recognition tasks, its key component of visual complexity is necessary for many real-world decision-making problems, such as self-driving cars, household robots, and unmanned autonomous vehicles \cite{zhu2017target}. Additionally, advances in computer vision algorithms, specifically around convolutional neural networks, were the motivation for the pixel-to-control approach eventually found in the Deep-Q network \cite{mnih2015human}.

{\bf Physical Complexity} - 
Many of the applied tasks researchers are interested in solving with AI involve not only rich sensory information, but a rich control scheme in which agents can interact with their dynamic environments in complex ways \cite{bicchi2000robotic,levine2016end}. The need for complex interaction often comes with the need for environments which replicate the physical properties of the target domain, typically the real world. This realism is essential to problems where the goal is to transfer a policy learned within a simulator to the real world, as would be the case for most robotics applications \cite{rusu2016sim,tobin2017domain,andrychowicz2018learning}. 

{\bf Task Logic Complexity} - 
A third axis is the complexity of the tasks defined within the environment. The game of Go, for example, which has long served as a test-bed for AI research, contains neither complex visuals nor complex physical interactions. Rather, the complexity comes from the large search space of possibilities open to the agent at any given time, and the difficulty in evaluating the value of a given board configuration \cite{muller2002computer,silver2017mastering}. Meaningful simulation platforms should enable designers to naturally create such problems for the learning agents within them. These complex tasks might display hierarchical structure, a hallmark of human intelligence \cite{botvinick2008hierarchical}, or vary from instance to instance, thus requiring meta-learning or generalization to solve \cite{wang2016learning}. The tasks may also be presented in a sequential manner, where independent sampling from a fixed distribution is not possible. This is often the case for human task acquisition in the real world, and the ability to learn new tasks over time is seen as a key-component of continual learning \cite{ring1994continual}, and ultimately systems capable of artificial general intelligence \cite{schmidhuber2015learning,schmidhuber2018one}. 

{\bf Social Complexity} - 
The acquisition of complex skills via learning in mammals is believed to have evolved hand-in-hand with their ability to hold relationships within their social groups \cite{arbib2008primate}. At least one strong example of this exists within the human species, with language primarily being the development of a tool for communication in a social setting. As such, the development of social behavior among groups of agents is of particular interest to many researchers in the field of AI. There are also classes of complex behavior which can only be carried out at the population level, such as the coordination needed to build modern cities~\cite{baker2019emergent}. Additionally, the ability for multiple species to interact with one another is a hallmark of the development of ecosystems in the world, and would be desirable to simulate as well. A simulation platform designed to allow the study of communication and social behavior should then provide a robust multi-agent framework which enables interaction between agents of both the same population as well as interaction between groups of agents drawn from separate distributions.

\subsection{Simulation Properties}
In addition to the properties above, there are practical constraints imposed by the simulator itself which must be taken into consideration when designing environments for experimentation. Specifically, simulated environments must be flexibly controlled by the researcher and must run in a fast and distributed manner in order to provide the iteration time required for experimental research.

{\bf Fast \& Distributed Simulation} - 
Depending on the sample efficiency of the method used, modern machine learning algorithms often require up to billions of samples in order to converge to an optimal solution \cite{espeholt2018impala,puigdomenech2020agent57}. As such, the ability to collect that data as quickly as possible is paramount. One of the most appealing properties of a simulation is the ability for it to be run at a speed often orders of magnitude greater than that of the physical world. In addition to this increase in speed, simulations can often be run in parallel, allowing for orders of magnitude greater data collection than real-time serial experience in the physical world. The faster such algorithms can be trained, the greater the speed of iteration and experimentation that can take place, leading to faster development of novel methods. 

{\bf Flexible Control} - 
A simulator must also allow the researcher or developer a flexible level of control over the configuration of the simulation itself, both during development and at runtime. While treating the simulation as a black-box has been sufficient for certain advances in recent years \cite{mnih2015human}, in many cases it also inhibits use of a number of advanced machine learning approaches in which more dynamic feedback between the training process and the agents is essential. Curriculum learning  \cite{bengio2009curriculum}, for example, entails initially providing a simplified version of a task to an agent, and slowly increasing the task complexity as the agent’s performance increases \cite{bengio2009curriculum}. This method was used to achieve near human-level performance in a recent VizDoom competition \cite{wu2016training}. Such approaches are predicated on the assumption that the user has the capacity to alter the simulation to create such curricula in the first place.
Additionally, domain randomization \cite{tobin2017domain} involves introducing enough variability into the simulation so that the models learned within the simulation can generalize to the real world. This often works by ensuring that the data distribution of the real world is covered within all of the variations presented within the simulation \cite{tobin2017domain}. This variation is especially important if the agent depends on visual properties of the environment to perform its task. It is often the case that without domain randomization, models trained in simulation suffer from a “reality gap” and perform poorly. Concretely, performing domain randomization often involves dynamically manipulating textures, lighting, physics, and object placement within a scene.

\section{A Survey of Existing Simulators}

When surveying the landscape of simulators, environments, and platforms, we find that there exist four categories into which these items can be organized. 

(1) The first is \textit{Environment} which consists of single, fixed environments that act as black-boxes from the perspective of the agent. Examples of these include the canonical CartPole or MountainCar tasks \cite{sutton2018reinforcement}, a single game from the ALE, such as Pitfall! \cite{bellemare2013arcade}, CoinRun \cite{cobbe2019quantifying}, and the Obstacle Tower environment \cite{juliani2019obstacle}. 

(2) The second is \textit{Environment Suite}. These consist of sets of environments packaged together and are typically used to benchmark the performance of an algorithm or method along some dimensions of interest. In most cases these environments all share the same or similar observation and action spaces, and require similar, but not necessarily identical skills to solve. Examples of this include the ALE \cite{bellemare2013arcade}, DMLab-30 \cite{espeholt2018impala}, the Hard Eight \cite{gulcehre2019r2d3}, AI2Thor \cite{kolve2017ai2}, OpenAI Retro \cite{nichol2018retro}, DMControl \cite{deepmindcontrolsuite2018}, and ProcGen \cite{cobbe2019leveraging}. 

(3) The third category is \textit{Domain-specific Platform}. This describes platforms which allow the creation of sets of tasks within a specific domain such as locomotion or first-person navigation. These platforms are distinguished from the final category by their narrow focus in environments types. This can include limitations to the perspective the agent can take, the physical properties of the environment, or the nature of the interactions and tasks possible within the environment. Examples of this category include Project Malmo \cite{johnson2016malmo}, VizDoom \cite{kempka2016vizdoom}, Habitat \cite{habitat19iccv}, DeepMind Lab \cite{beattie2016deepmind}, PyBullet \cite{coumans2016pybullet} and GVGAI \cite{perez2016general}. 

(4) The fourth and final category is the \textit{General Platform} whose members are capable of creating environments with arbitrarily complex visuals, physical and social interactions, and tasks. The set of environments that can be created by platforms in this category is a super-set of those that can be created by or are contained within the other three categories. In principle, members of these categories can be used to define any AI research environment of potential interest. We find that modern video game engines represent a strong candidate for this category. In particular, we propose the Unity engine along with a toolkit for AI interactions such as ML-Agents as an example of this category. Note that  other game engines such as the Unreal engine could serve as general platforms for AI research. The important missing element however is the set of useful abstractions and interfaces for conducting AI research, something present in all examples listed here, but not inherently part of any given game engine or programming language. See Table~\ref{table:taxonomy} for a representative set of examples of the environments and platforms within this taxonomy. 

\begin{table}[h!]
\centering
\begin{tabular*}{\textwidth}{c@{\extracolsep{\fill}}cccc}
 \hline
 \textbf{Single Env} & \textbf{Env Suite} & \textbf{Domain-Specific Platform} & \textbf{General Platform}\\
 \hline
 Cart Pole & ALE & MuJoCo & Unity \& ML-Agents \\ 
 Mountain Car & DMLab-30 & DeepMind Lab & \\
 Obstacle Tower & Hard Eight & Project Malmo & \\
 Pitfall! & AI2Thor & VizDoom & \\
 CoinRun & OpenAI Retro & GVGAI & \\
 Ant & DMControl & PyBullet & \\
  & ProcGen & & \\
 \hline
\end{tabular*}
\caption{Taxonomy of simulators based on flexibility of environment specification. Includes a subset of examples for illustrative purposes.}
\label{table:taxonomy}
\end{table}

\subsection{Common Simulators}

In recent years, there have been a number of simulation platforms developed for the purpose of providing challenges and benchmarks for deep reinforcement learning algorithms. Many of these platforms are based on existing games or game engines and carry with them specific strengths and weaknesses. While not exhaustive of all currently available platforms, below we survey a few of the simulators described in the previous section, taking examples from the middle two categories. 

\subsubsection{Arcade Learning Environment}

The release of the Arcade Learning Environment (ALE) contributed to much of the recent resurgence of interest in reinforcement learning. This was thanks to the development of the Deep Q-Network, which was able to achieve superhuman level performance on dozens of emulated Atari console games within the ALE by learning only from pixel inputs \cite{mnih2015human}. The ALE provides a Python interface for launching and controlling simulations of a few dozen Atari 2600 games. As such, the ALE falls into the category of environment suite. When considering the simulation criteria described above, the ALE provides visual complexity through pixel-based rendering, task-logic complexity in the form of hierarchical problems within some games such as Montezuma’s Revenge, and high-performance simulation with an emulation able to run at thousands of frames per second \cite{bellemare2013arcade}. Its downsides include deterministic environments, relatively simple visuals, a lack of realistic physics, single-agent control, and a lack of flexible control of the simulation configuration. In general, once an environment that is part of the ALE is launched, it is immutable and a complete black box from the perspective of the agent. Furthermore, all of the environments provided in the ALE have been solved with greater than human performance \cite{puigdomenech2020agent57}. As such, there is little room for meaningful improvement over the state of the art with the exception of the domain of few-shot learning. This is apparent in the fact that even Agent57, the current state of the art algorithm takes orders of magnitude more training time than humans on a large number of the environments in the ALE. 

\subsubsection{DeepMind Lab}

Built from the Quake III game engine, DeepMind Lab (Lab) was released in 2016 as the external version of the research platform used by DeepMind \cite{beattie2016deepmind}. Designed in the wake of public adoption of the ALE, Lab contains a number of features designed to address the other platform’s shortcomings. By using a 3D game-engine, complex navigation tasks similar to those studied in robotics and animal psychology could be created and studied within Lab \cite{leibo2018psychlab}. The ability to create a set of specific kinds of tasks makes DeepMind Lab a domain-specific platform. The platform contains primitive physics enabling a level of prediction about the quality of the world and allows researchers to define their own environmental variations. Additionally, it allows for basic multi-agent interactions using language \cite{espeholt2018impala}. The limitations of this platform, however, are largely tied to the dated nature of the underlying rendering and physics engine, which was built using decades-old technology. As such, the gap in quality between the physical world and the simulation provided via Lab is relatively large. Furthermore, the engine was also designed to enable first-person shooter games and so the environments built using Lab are limited to agents with a first-person perspective. 

\subsubsection{Project Malmo}

Another popular simulation platform is Project Malmo (Malmo)~\cite{johnson2016malmo}. Based on the exploration and building game Minecraft, the platform provides a large amount of flexibility in defining scenarios and environment types making it a domain-specific platform. As a result, there have been a number of research projects exploring multi-agent communication, hierarchical control, and planning using the platform \cite{oh2016control,shu2017hierarchical,tessler2017deep}. The limitations of the platform, however, are bound tightly with the underlying limitations of the Minecraft engine itself. Due to the low-polygon pixelated visuals, as well as the rudimentary physics system, Minecraft lacks both the visual as well as the physical complexity that is desirable from a modern platform. The platform is also limited to describing scenarios which are only possible within the logic of Minecraft.%, making Malmo a domain-specific platform.

\subsubsection{Physics Simulators}

The MuJoCo physics engine has become a popular simulation platform for benchmarking model-free continuous control tasks, thanks to a set of standard tasks built on top of MuJoCo, provided with OpenAI Gym and the DeepMind Control Suite \cite{todorov2012mujoco,brockman2016openai,tassa2018deepmind}. High quality physics simulation combined with a number of standardized benchmarks has led to the platform being the primary choice for researchers interested in examining the performance of continuous control algorithms. The nature of the MuJoCo engine, however, poses limitations for more general AI research. The first is around the limited visual rendering capabilities of the engine, preventing the use of complex lighting, textures, and shaders. The second is the restrictions of the physics engine itself and that MuJoCo models are compiled which makes more difficult the creation of dynamic “game-like” environments, where many different objects would be instantiated and destroyed in real-time during simulation. More dynamic environments are often necessary to pose tasks which require greater planning or coordination to solve. %As such, MuJoCo is limited to being a domain-specific platform, with DeepMind Control Suite being an example of an environment suite.
The PyBullet physics engine has also been used as a platform to study deep reinforcement learning algorithms as well as sim-to-real \cite{coumans2016pybullet,tan2018sim}. Similar to MuJoCo, the PyBullet simulator lacks the ability to provide high-fidelity visuals, and the nature of the physics engine limits the scope of possible tasks to be defined.%, making it also a domain-specific platform.

\subsubsection{VizDoom}

Based on the game Doom, VizDoom provides researchers with the ability to create tasks which involve first-person navigation and control \cite{kempka2016vizdoom}. Through a 2017 AI deathmatch competition, the platform enabled the development of a number of compelling approaches to Deep Reinforcement Learning, including utilizing learning curricula \cite{wu2016training}, novel algorithm design \cite{dosovitskiy2016learning}, and memory systems \cite{lample2017playing}. Like DeepMind Lab, the platform is mainly restricted by the underlying game engine, which was built for a decades-old first-person shooter game. As such, the visual and physical complexity possible in the environments created using VizDoom are relatively limited. It also is restricted to simulating artificial agents with only a first-person perspective.%, making the platform domain-specific. 

\section{The Unity Platform}\label{UnityPlatform}

Unity is a real-time 3D development platform that consists of a rendering and physics engine as well as a graphical user interface called the Unity Editor. Unity has received widespread adoption in the gaming, AEC (Architecture, Engineering, Construction), auto, and film industries and is used by a large community of game developers to make a variety of interactive simulations, ranging from small mobile and browser-based games to high-budget console games and AR/VR experiences.

Unity's historical focus on developing a general-purpose engine to support a variety of platforms, developer experience levels, and game types makes the Unity engine an ideal candidate simulation platform for AI research.
The flexibility of the underlying engine enables the creation of tasks ranging from simple 2D gridworld problems to complex 3D strategy games, physics-based puzzles, or multi-agent competitive games possible. Unlike many of the research platforms discussed above, the underlying engine is not restricted to any specific genre of gameplay or simulation, making Unity a general platform. Furthermore, the Unity Editor enables rapid prototyping and development of games and simulated environments. 

A Unity Project consists of a collection of Assets. These typically correspond to files within the Project. Scenes are a special type of Asset which define the environment or level of a Project. Scenes contain a definition of a hierarchical composition of GameObjects, which correspond to the actual objects (either physical or purely logical) within the environment. The behavior and function of each GameObject is determined by the components attached to it. There are a variety of built-in components provided with the Unity Editor, including Cameras, Meshes, Renderers, RigidBodies, and many others. It is also possible to define custom components using C\texttt{\#} scripts or external plugins. 

\subsection{Engine Properties}

This section examines the properties of the Unity engine from the perspectives described in Section~\ref{simulator_properties}. We demonstrate that Unity enables the complexity necessary along the key dimensions of environment properties for the creation of challenging learning environments.

\subsubsection{Environment Properties}

\textbf{Sensory Complexity} - The Unity engine enables high-fidelity graphical rendering. It supports pre-baked as well as real-time lighting and the ability to define custom shaders, either programmatically or via a visual scripting language. As such, it is possible to quickly render near-photorealistic imagery to be used as training data for a machine learning model. It is also possible to render depth information, object masks, infrared, or images with noise injected into it through the use of custom shaders. Furthermore, the engine provides a means of defining audio signals which can serve as potential additional observational information to learning agents, as well as ray-cast based detection systems which can simulate Lidar.

\noindent
\textbf{Physical Complexity} -  Physical phenomena in Unity environments can be simulated with either the Nvidia PhysX or Havok Physics engines. This enables research in environments with simulated rigid body, soft body, particle, and fluid dynamics as well as ragdoll physics. Furthermore, the extensible nature of the platform enables the use of additional 3rd party physics engines if desired. For example, there are plugins available for Unity which provide both the Bullet and MuJoCo physics engines as alternatives to PhysX \footnote{\url{https://assetstore.unity.com/packages/tools/physics/bullet-physics-for-unity-62991}; \url{http://www.mujoco.org/book/unity.html}}.

\noindent
\textbf{Task Logic Complexity} - The Unity Engine provides a rich and flexible scripting system via C\texttt{\#}. This system enables any form of gameplay or simulation to be defined and dynamically controlled. In addition to the scripting language, the GameObject and component system enables managing multiple instances of agents, policies, and environments, making it possible to define complex hierarchical tasks, or tasks which would require meta-learning to solve. 

\noindent
\textbf{Social Complexity} - The nature of the Unity scripting language and component system makes the posing of multi-agent scenarios simple and straightforward. Indeed, because the platform was designed to support the development of multi-player video games, a number of useful abstractions are already provided out of the box, such as the Multiplayer Networking system\footnote{\url{https://unity3d.com/learn/tutorials/s/multiplayer-networking}}.  

\subsubsection{Simulation Properties}

\textbf{Fast \& Distributed Simulation} - The physics and frame rendering of the Unity engine take place asynchronously. As such, it is possible to greatly increase the speed of the underlying simulation without the need to increase the frame rate of the rendering process. It is also possible to run Unity simulations without rendering if it is not critical to the simulation. In scenarios where rendering is desirable, such as learning from pixels, it is possible to control the frame rate and speed of game logic. Extensive control of the rendering quality also makes it possible to greatly increase the frame rate when desired. 
The added capabilities of the Unity engine do add additional overhead when attempting to simulate in a large-scale distributed fashion. The memory footprint of a Unity simulation is also larger than that of environments from other platforms such as an Atari game in the ALE.

\noindent
\textbf{Flexible Control} - It is possible to control most aspects of the simulation programmatically, enabling researchers to define curricula, adversarial scenarios, or other complex methods of changing the learning environment during the training process. For example, GameObjects can be conditionally created and destroyed in real-time. In Section~\ref{core_functionality}, we discuss ways in which further control of the simulation is made possible via exposed simulation parameters and a Python API. 

\subsection{Unity Editor and Services}

\begin{figure}[h]
    \centering
    \includegraphics[width=1.0\textwidth]{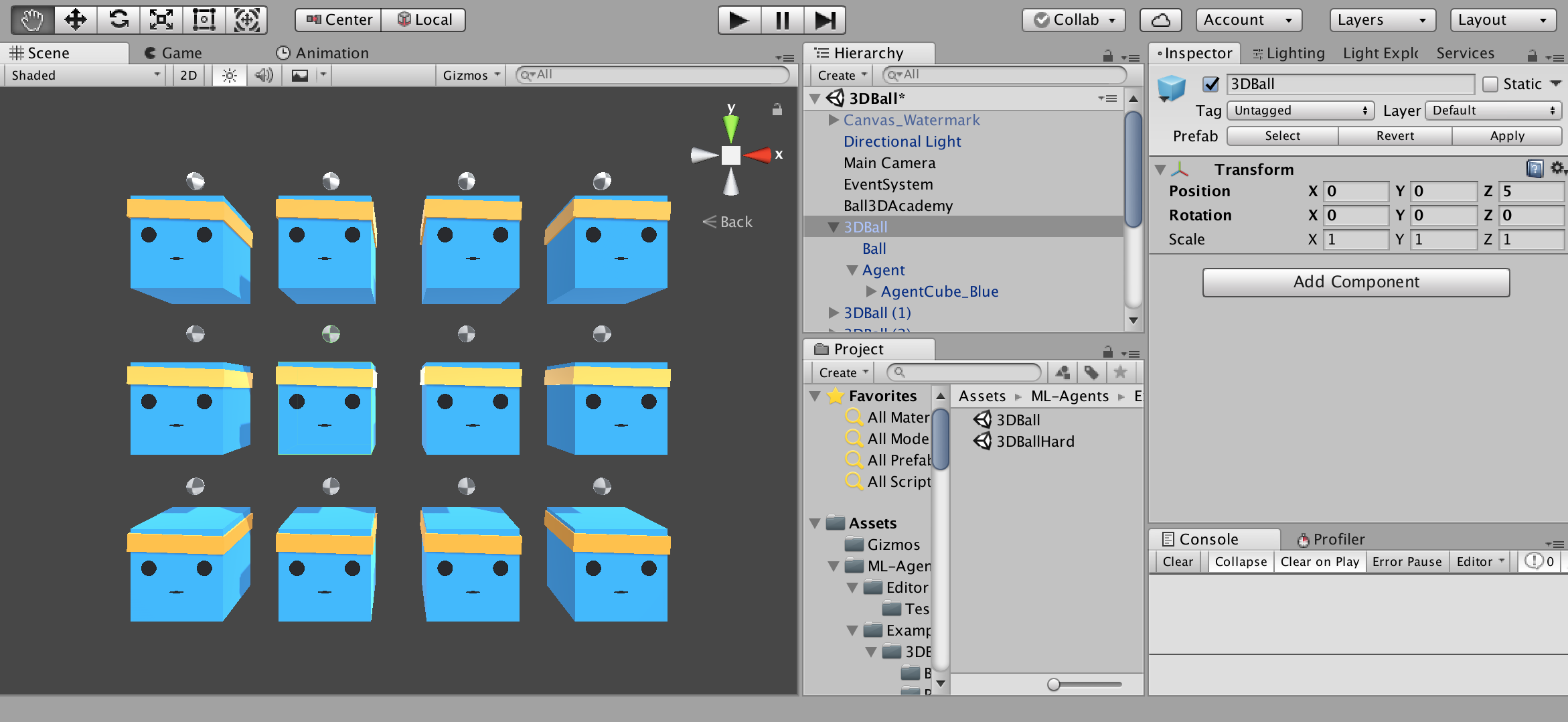}
    \caption{The Unity Editor window on macOS.}
    \label{fig:editor}
\end{figure}

The Unity Editor (Figure~\ref{fig:editor}) is a graphical user interface used to create the content for 2D, 3D and AR / VR experiences. It is available on Windows, Mac and Linux.

The Unity Editor and its services provide additional benefits for AI research:

\begin{enumerate}[leftmargin=0.5cm]
\item \emph{Create custom Scenes} - Unity provides a large number of guides and tutorials on how to create Scenes within the Editor. This enables developers to quickly experiment with new environments of varying complexities, or novel tasks. Furthermore, an online asset store which contains tens of thousands of free and paid assets provides users access to a huge diversity of pre-built entities for their scene. 
\item \emph{Record local, expert demonstrations} - The Unity Editor includes a Play mode which enables a developer to begin a simulation and control one or more of the agents in the Scene via a keyboard or game controller. This can be used for generating expert demonstrations to train and evaluate imitation learning (IL) algorithms. 
\item \emph{Record large-scale demonstrations} - One the most powerful features of the Unity Editor is the ability to build a Scene to run on more than 20 platforms ranging from wearables to mobile and consoles. This enables developers to distribute their Scenes to a large number of devices (either privately or publicly through stores such as the Apple App Store or Google Play). This can facilitate recording expert demonstrations from a large number of experts or measuring human-level performance from a user (or player) population.
\end{enumerate}

\section{The Unity ML-Agents Toolkit}\label{core_functionality}

The Unity ML-Agents Toolkit\footnote{This describes version 1.0; the most recent release at the time of writing.} is an open source project which enables researchers and developers to create simulated environments using the Unity Editor and interact with them via a Python API. The toolkit provides the ML-Agents SDK which contains all functionality necessary to define environments within the Unity Editor along with the core C\texttt{\#} scripts to build a learning pipeline. 

The features of the toolkit include a set of example environments, state of the art RL algorithms Soft Actor-Critic (SAC)~\cite{haarnoja2018soft} and Proximal Policy Optimization (PPO)~\cite{schulman2017proximal}, the IL algorithms Generative Adversarial Imitation Learning (GAIL)~\cite{ho2016generative} and Behavioral Cloning (BC)~\cite{hussein2017imitation}, support for Self-Play~\cite{baker2019emergent,bansal2017emergent} in both symmetric and asymmetric games, as well as the option to extend algorithms and policies with the Intrinsic Curiosity Module (ICM) \cite{pathak2017curiosity} and Long-Short-Term Cell (LSTM) \cite{hochreiter1997long}, respectively. As the platform grows, we intend to provide additional algorithms and model types.
In what follows, we outline the key components of the toolkit as well as provide benchmark results with SAC and PPO on the Unity example environments.

\begin{figure}[h]
    \centering
    \includegraphics[width=0.8\textwidth]{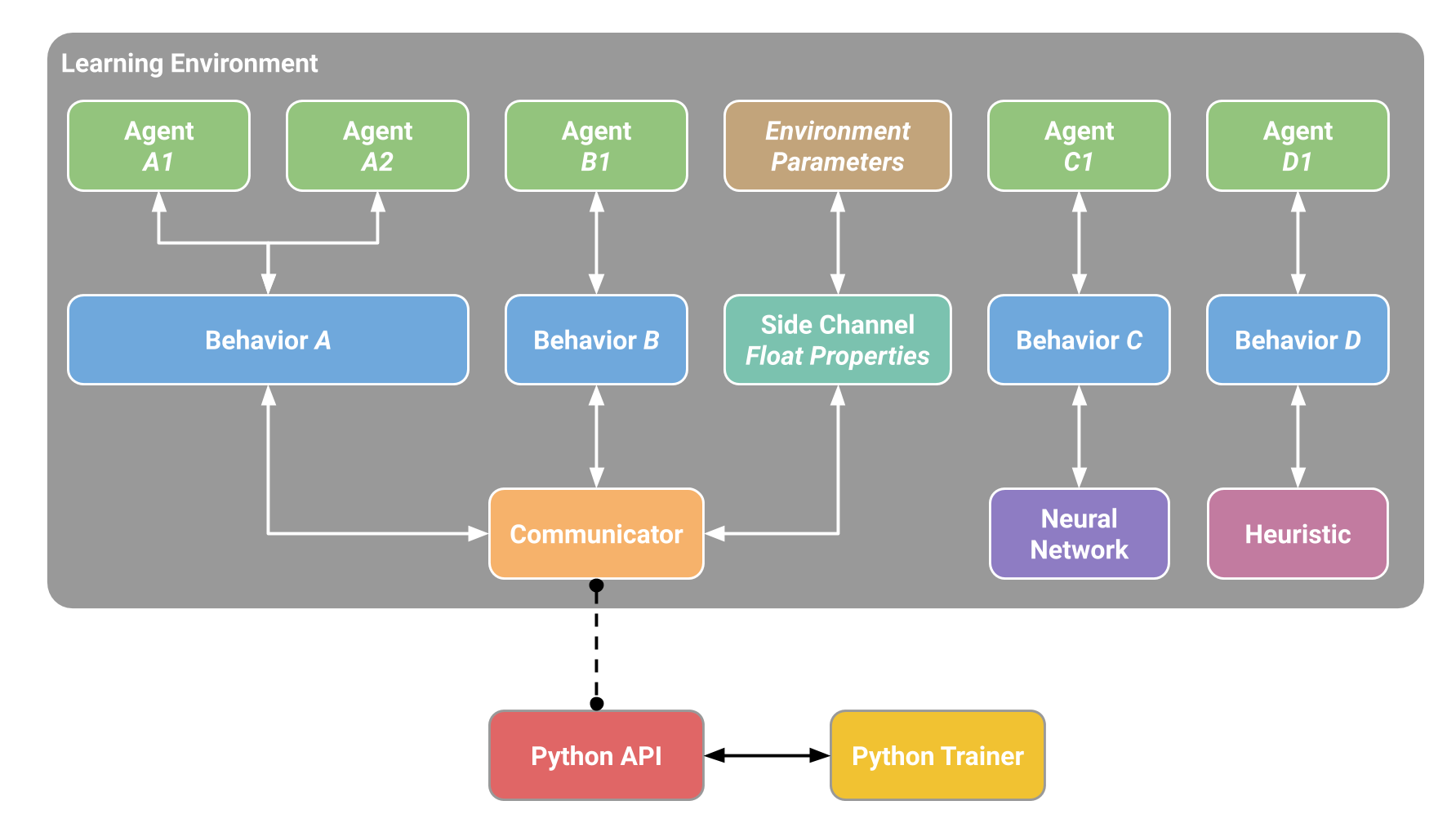}
    \caption{A Learning Environment (as of version 1.0) created using the Unity Editor contains Agents and an Academy. The Agents are responsible for collecting observations and executing actions. The Academy is responsible for global coordination of the environment simulation.}
    \label{fig:diagram}
\end{figure}

\subsection{ML-Agents SDK}

The three core entities in the ML-Agents SDK are Sensors, Agents, and an  Academy. The Agent component is used to directly indicate that a GameObject within a scene is an Agent, and can thus collect observations, take actions, and receive rewards. The agent can collect observations using a variety of possible sensors corresponding to different forms of information such as rendered images, ray-cast results, or arbitrary length vectors. Each Agent component contains a policy labeled with a {\it behavior name}.

Any number of agents can have a policy with the same behavior name. These agents will execute the same policy and share experience data during training. Additionally, there can be any number of behavior names for policies within a scene enabling simple construction of multi-agent scenarios with groups or individual agents executing many different behavior types.  A policy can reference various decision-making mechanisms including player input, hard-coded scripts, internally embedded neural network models, or via interaction through the Python API. It is possible for agents to ask for decisions from policies either at a fixed or dynamic interval, as defined by the developer of the environment.

The reward function, used to provide a learning signal to the agent, can be defined or modified at any time during the simulation using the Unity scripting system.  Likewise, simulation can be placed into a done state either at the level of an individual agent or the environment as a whole. This happens either via a Unity script call or by reaching a predefined max step count.  

The Academy is a singleton within the simulation, and is used to keep track of the steps of the simulation and manage the agents. The Academy also contains the ability to define environment parameters, which can be used to change the configuration of the environment at runtime. Specifically, aspects of environmental physics and textures, sizes and the existence of GameObjects are controlled via exposed parameters which can be re-sampled and altered throughout training. For example, the gravity in the environment can fluctuate every fixed interval or additional obstacles can spawn when an agent reaches a certain proficiency.  This enables evaluation of an agent on a train/test split of environment variations and facilitates the creation of curriculum learning scenarios~\cite{bengio2009curriculum}.

\subsection{Python Package}

The provided Python package\footnote{\url{https://pypi.org/project/mlagents/}} contains a class called UnityEnvironment that can be used to launch and interface with Unity executables (as well as the Editor) which contain the required components described above. Communication between Python and Unity takes place via a gRPC communication protocol, and utilizes protobuf messages.

We also provide a set of wrapper APIs, which can be used to communicate with and control Unity learning environments through the standard gym interface used by many researchers and algorithm developers \cite{brockman2016openai}. These gym wrappers enable researchers to more easily swap in Unity environments to an existing reinforcement learning system already designed around the gym interface.

\subsection{Performance Metrics}

It is essential that an environment be able to provide greater than real-time simulation speed. It is possible to increase Unity ML-Agents simulations up to one hundred times real-time. The possible speed increase in practice, however, will vary based on the computational resources available, as well as the complexity of the environment. In the Unity Engine, game logic, including physics, can be run independently from the rendering of frames. As such, environments which do not rely on visual observations, such as those that use ray-casts for example, can benefit from simulation at speeds greater than those that do. See Table~\ref{table:data} for performance metrics when controlling environments from the Python API. 

 \begin{table}[h!]
 \centering
 \begin{tabular}{|c c c c c|} 
  \hline
  \textbf{Environment} & \textbf{Observation Type} & \textbf{\# Agents} & \textbf{Mean (ms)} & \textbf{Std (ms)} \\ [0.5ex] 
  \hline
  Basic & Vector(1) & 1 & 0.803 & 0.005 \\ 
  3D Ball & Vector(8) & 12 & 5.05 & 0.039 \\
  GridWorld & Visual(84x84x3) & 1 & 2.04 & 0.038 \\
  Visual Food Collector & Visual(84x84x3) & 4 & 9.23 & 0.556 \\ [1ex] 
  \hline
 \end{tabular}
 \caption{Performance benchmark when using the Python API to control a Learning Environment from the same machine by calling $env.step()$. Mean and standard deviation in time averaged over 1000 simulation steps.}
 \label{table:data}
 \end{table}

\subsection{Example Environments}

\begin{figure}[h]
    \centering
    \includegraphics[width=1.0\textwidth]{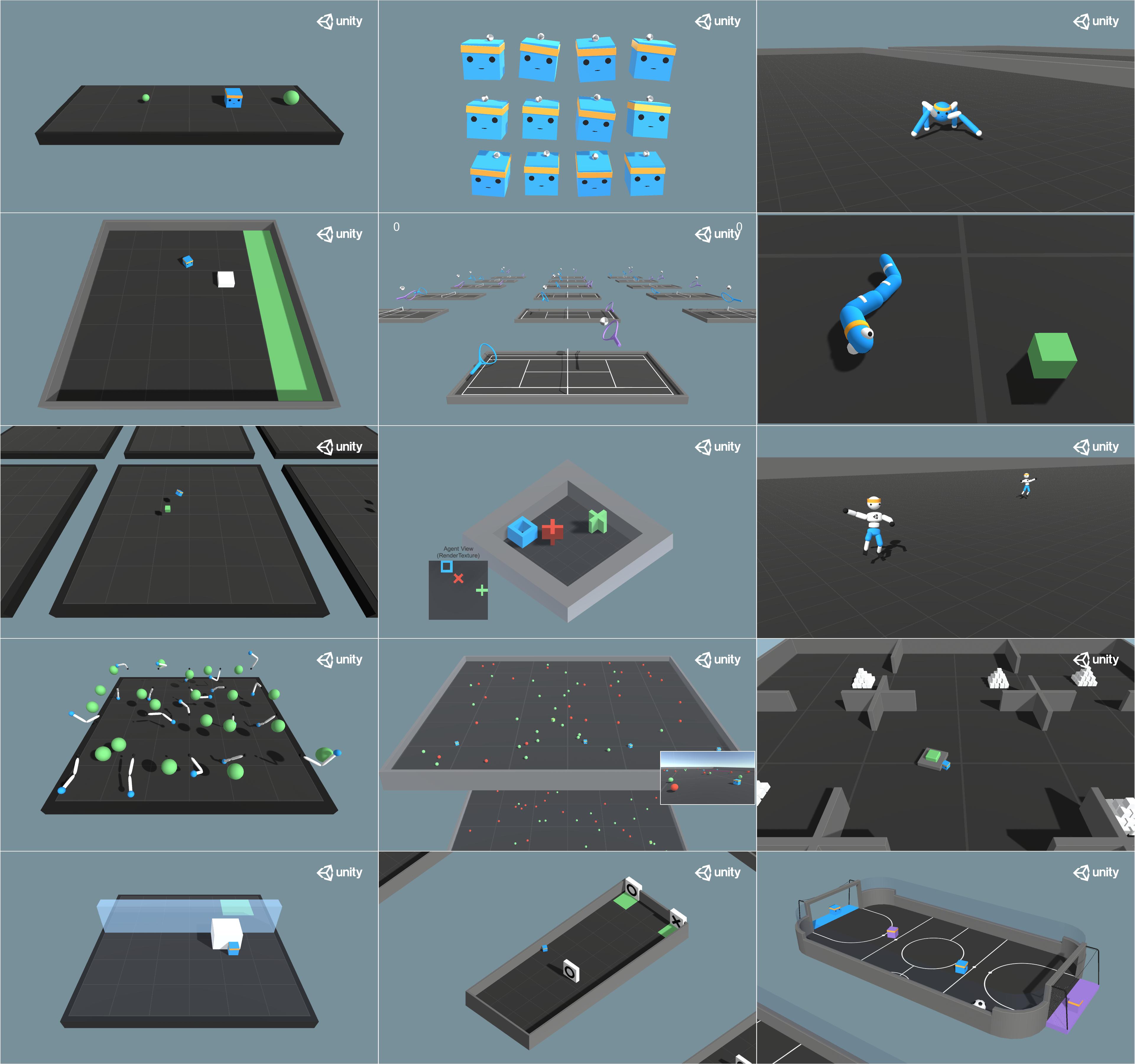}
    \caption{Images of the fourteen included example environments as of the v0.11 release of the Unity ML-Agents Toolkit. From Left-to-right, up-to-down: (a) Basic, (b) 3DBall, (c) Crawler, (d) Push Block, (e) Tennis, (f) Worm, (g) Bouncer, (h) Grid World, (i) Walker, (j) Reacher, (k) Food Collector,   (l) Pyramids,  (m) Wall Jump, (n) Hallway, (o) Soccer Twos.}
    \label{fig:environments}
\end{figure}

The Unity ML-Agents Toolkit contains a number of example environments in addition to the core functionality. These environments are designed to both be usable for benchmarking RL and IL algorithms as well as templates to develop novel environments and tasks. These environments contain examples of single and multi-agent scenarios, with agents using either vector or visual observations, taking either discrete or continuous actions, and receiving either dense or sparse rewards.
See Figure~\ref{fig:environments} for images of the included example environments and below for environment descriptions.

\begin{enumerate}[leftmargin=0.5cm,itemindent=0.0cm,label=(\alph*)]
\item \textit{Basic} - A linear movement task where the agent (blue cube) must move left or right to rewarding states. The goal is to move to the most rewarding state.
\item \textit{3D Ball} - A balance-ball task where the agent controls the rotation of the platform. The goal is to balance the platform in order to keep the ball on it for as long as possible.
\item \textit{Crawler} - Physics-based creatures with 4 arms and 4 forearms. The goal is to move toward the goal direction as quickly as possible without falling.
\item \textit{Push Block} - A platforming environment where the agent can push a block around. The goal is to push the block to the target area (black white grid).
\item \textit{Tennis} - Two-player game where agents control rackets to bounce ball over a net. The goal is to bounce ball to the other side instead of dropping the ball or sending ball out of bounds.
\item \textit{Worm} - A physics-based three joint locomotion agent which must move toward a goal location as quickly as possible. 
\item \textit{Bouncer} - A bouncing task where the agent (blue cube) can jump with a certain speed and angle when it touches the ground. The goal is to catch the floating food object with as few jumps as possible. 
\item \textit{Grid World} - A version of the classic grid-world task. Scene contains agent (blue square), target, and obstacles. The goal is to navigate to the target while avoiding the obstacles.
\item \textit{Walker} - Physics-based humanoids with 26 degrees of freedom of its body-parts. The goal is to move toward the goal direction as quickly as possible without falling.
\item \textit{Reacher} - Double-jointed arm which can move to target locations. The goal is to move its hand to the target location (green sphere), and keep it there.
\item \textit{Food Collector} - A multi-agent environment where agents (blue cube) compete to collect bananas. The goal is to move to as many yellow bananas as possible while avoiding blue bananas.
\item \textit{Pyramids} - Environment where the agent (blue cube) needs to press a button to spawn a pyramid, then navigate to the pyramid, knock it over, and move to the gold brick at the top. The goal is to move to the golden brick on top of the spawned pyramid.
\item \textit{Wall Jump} - A platforming environment with a wall and a yellow block that can be pushed around, and an agent (blue cube) that can move, rotate and jump. The goal is to reach the target (white black grid) on the other side of the wall. If the wall is too high, the agent sometimes needs to push the white block near the wall, jump onto it to reach its target. The agent trains two policies---one for big walls (requires the small block) and one for small walls. 
\item \textit{Hallway} - Environment where the agent (blue cube) needs to find information in a room, remember it, and use it to move to the correct target. The goal is to move to the target (black white grid) which corresponds to the color of the block in the room.
\item \textit{Soccer} - Environment where four agents compete in a 2 vs 2 toy soccer game. All agents are equal and tasked with keeping the ball out of their own goal and scoring in the opponents. 
\item \textit{StrikersVsGoalie} - A soccer variant with three agents of two different kinds in the environment; two Strikers and one Goalie. The goal of the Striker agents is to push the ball into the goal area while the Goalie tries to prevent the ball from entering its own goal area.
\end{enumerate}

\begin{figure}[ht]
    \centering
    \includegraphics[width=0.9\textwidth]{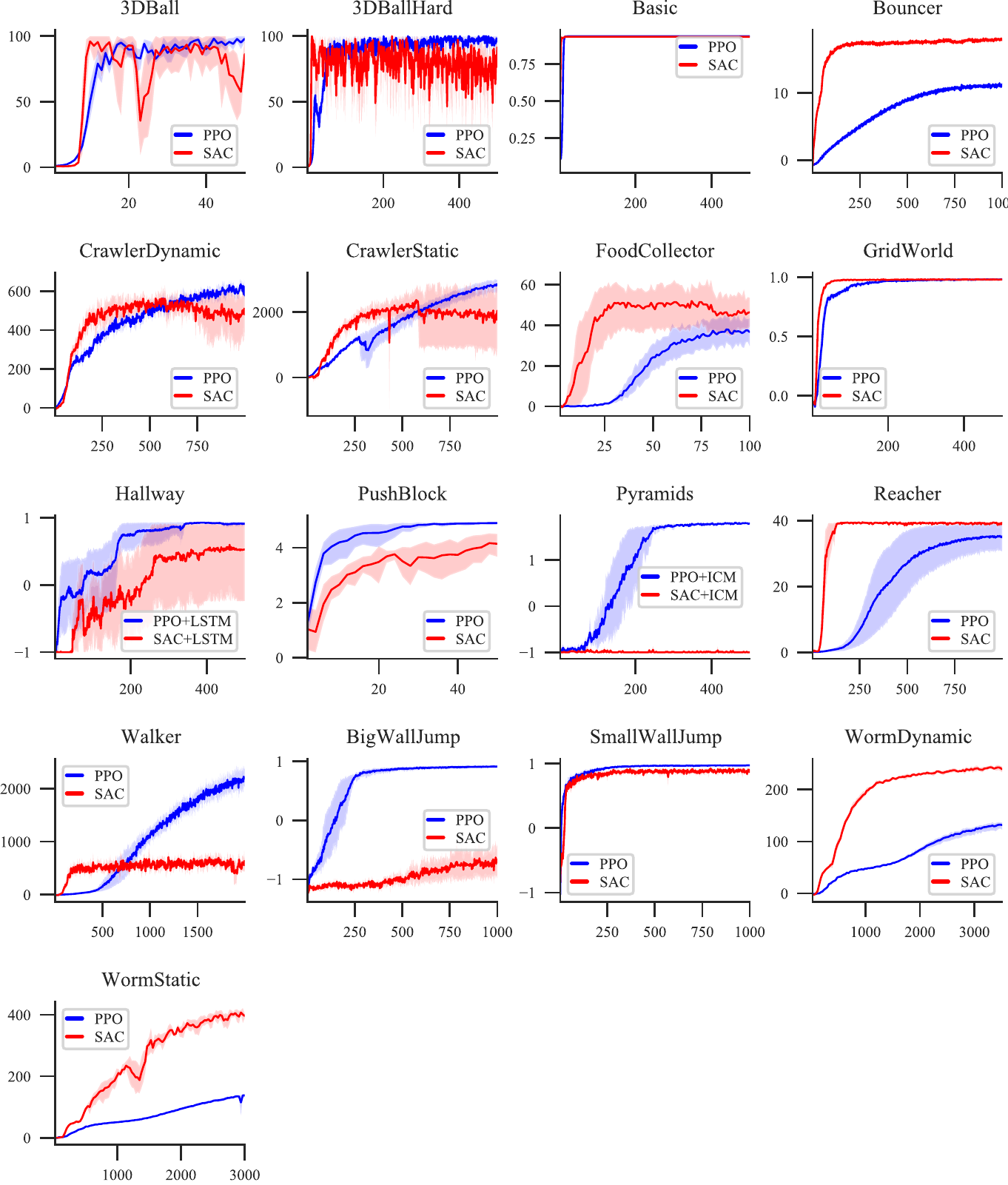}
    %\scalebox{.925}{\input{images/results.pgf}}
    %\includegraphics[width=1.1\textwidth,center]{images/results-15.png}
    \caption{Mean cumulative episodic reward (y-axis) over simulation time-steps (in thousands, x-axis) during training and evaluation. We compare PPO (blue line) and SAC (red line) performances. Results presented are based on five separate runs, with a 95\% confidence interval. LSTM indicates an LSTM unit is used in the network. ICM indicates the Intrinsic Curiosity Module is used during training.}
    \label{fig:results}
\end{figure}

\begin{figure}[ht]
    \centering
    \includegraphics[width=0.9\textwidth]{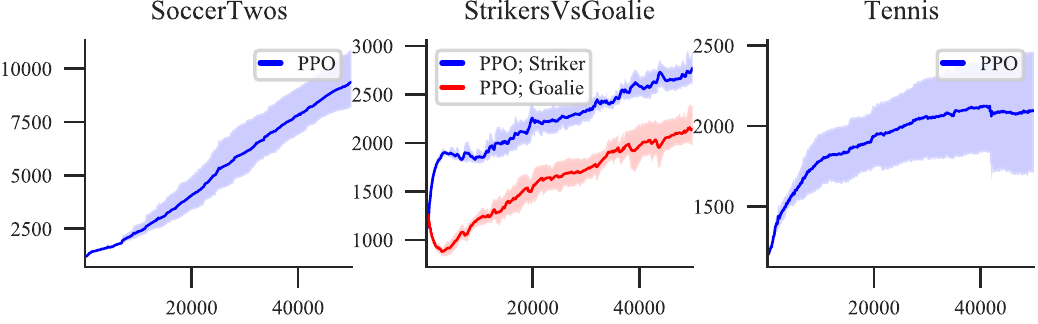}
    \caption{Mean episodic ELO (y-axis) over simulation time-steps (in thousands, x-axis) during training with Self-Play and PPO. In symmetric environments, the ELO of the learning policy is plotted (blue line) and in asymmetric environments (blue and red line) the ELO of both learning policies are plotted. Results presented are based on five separate runs, with a 95\% confidence interval.}
    \label{fig:self-play-results}
\end{figure}

For more information on the specifics of each of the environments, including the observations, actions, and reward functions, see the GitHub documentation\footnote{\url{https://github.com/Unity-Technologies/ml-agents/blob/master/docs/Learning-Environment-Examples.md}}.  Trained model files as well as hyperparameter specifications for replicating all of our results on the example environments are provided with the toolkit. See Figures~\ref{fig:results} and~\ref{fig:self-play-results}  below for baseline results on each example environment. These results describe the mean cumulative reward per-episode over five runs using PPO and SAC (plus relevant modifications).

%The toolkit also provides trained model files as well as hyperparameter specifications for replicating all of our results on the example environments provided with the toolkit. See Figure~\ref{fig:results} below for baseline results on each of our provided example environments. These results describe the mean cumulative reward per-episode over five runs using PPO and SAC (plus relevant modifications).% While SAC and PPO are able to consistently solve most environments, they do poorly on a few environments such as VisualPyramids, VisualPushBlock, VisualHallway, and SoccerTwos. We encourage researchers to develop solutions which achieve greater mean rewards or learn more quickly than the results provided here.

\begin{table}[ht]
\centering
\begin{tabular}{|p{3.75cm}p{1.75cm}p{1.75cm}p{1.75cm}p{1.75cm}|}
\hline
\textbf{Environment} & \textbf{PPO (mean)} & \textbf{PPO (std)} & \textbf{SAC (mean)} & \textbf{SAC (std)} \\ %\textbf{Random} & \textbf{Human (mean)} & \textbf{Human (std)} \\
\hline
3DBall & 98.03 & 2.95 & 86.36 & 12.08 \\
3DBallHard & 96.05 & 7.91 & 91.36 & 8.91 \\
Basic & 0.94 & 0.0 & 0.94 & 0.0 \\
Bouncer & 11.33 & 0.07 & 17.84 & 0.27 \\
CrawlerDynamic & 577.51 & 25.26 & 479.73 & 131.71 \\
CrawlerStatic & 2816.07 & 231.37 & 2042.97 & 1097.81 \\
FoodCollector & 36.6 & 8.42 & 46.43 & 7.93 \\
GridWorld & 0.98 & 0.0 & 0.98 & 0.0 \\
Hallway & 0.91 & 0.03 & 0.53 & 0.76 \\
PushBlock & 4.89 & 0.04 & 4.14 & 0.49 \\
Pyramids & 1.79 & 0.02 & -1.0 & 0.0 \\
Reacher & 35.28 & 4.43 & 39.29 & 0.11 \\
Walker & 2206.41 & 165.66 & 567.45 & 129.35 \\
BigWallJump & 0.91 & 0.02 & -0.66 & 0.29 \\
SmallWallJump & 0.97 & 0.0 & 0.89 & 0.04 \\
WormDynamic &  131.59 & 9.08 & 238.89 & 6.2\\
WormStatic & 152.54 & 4.02 &  396.26 &7.25\\
\hline
\end{tabular}
\caption{Table of cumulative episodic reward for the various example environments provided with the Unity ML-Agents Toolkit. Results are averaged over final score on five separate runs.}
\label{table:results}
\end{table}

\section{Research Using Unity and the Unity ML-Agents Toolkit}
In this section, we survey a collection of results from the literature which use Unity and/or the Unity ML-Agents Toolkit. The range of environments and algorithms reviewed here demonstrates the viability of Unity as a general platform. We also discuss the Obstacle Tower benchmark \cite{juliani2019obstacle} which serves as an example of the degree of environmental complexity achievable on Unity. The corresponding Obstacle Tower contest posed a significant challenge to the research community inspiring a number of creative solutions. We review the top performing algorithm to show the rallying effect a benchmark like this can have on innovation in the field.

\subsection{Domain-Specific Platforms and Algorithms}
The AI2Thor \cite{kolve2017ai2} platform provides a set of pre-built indoor scenes which are rendered using the Unity engine and a Python API for interacting with those environments using a first-person agent. Using the AI2Thor simulator, researchers demonstrated that it is possible to transfer a policy learned in simulation to a physical robot to complete an indoor-navigation task \cite{zhu2017target}. In the same vein, the Chalet platform uses Unity to provide a set of indoor navigation environments \cite{yan2018chalet}. Recent work at OpenAI has also taken advantage of the rendering capabilities of the Unity engine to aid in the development of a system used to transfer a robotic hand’s grasping policy from a simulator to a physical robot \cite{andrychowicz2018learning}. Unity has also been used to render a physical intersection in order to aid demonstration-based learning on real-world vehicles \cite{behbahani2019learning}. Finally, a set of benchmarks called ``Arena'' have been built using the Unity ML-Agents Toolkit which specifically focuses on multi-agent scenarios \cite{song2020arena}. The creators of Arena take care to note that Unity is selected over other engines and platforms for its generality.

Unity environments have been used in varied research such as intrinsic motivation~\cite{burda2018large,pathak2019self}, Neural Attention~\cite{ghani2018designing}, and semi-parametric reinforcement learning~\cite{jain2018semiparametric}. Of particular interest is the way in which Unity facilitated work which developed an algorithm for the morphological self-assembly of individually trained agents in order to achieve some higher order task like standing or locomotion~\cite{pathak2019morph}. The authors note that none of the standard benchmark environments support the co-evolution of control and morphology which {\it required them to create their own.} A general platform promotes experimentation with these types of highly original algorithms.

\subsection{Obstacle Tower}

The Obstacle Tower\footnote{https://github.com/Unity-Technologies/obstacle-tower-env} environment for deep reinforcement learning~\cite{juliani2019obstacle} demonstrates the extent of environmental complexity possible from the Unity platform. This benchmark uses procedural generation and sparse rewards in order to ensure that each instance of the task requires flexible decision-making. Each episode of Obstacle Tower consists of one-hundred randomly generated floors, each with an increasingly complex floor layout. Each floor layout is composed of rooms, which can contain puzzles, obstacles, enemies, or locomotion challenges. The goal of the agent is to reach the end room of each floor and to ascend to the top floor of the tower without entering a fail-state such as falling in a hole or being defeated by an enemy. This benchmark provided a significant challenge to contemporary RL algorithms, with baseline results showing test-time performance corresponding to solving on average five of 100 floors after 20 million time-steps of training. This is significantly worse than those of naive humans who have only interacted with the environment for five minutes and are able to solve on average 15 floors \cite{juliani2019obstacle}, and much worse than expert players who are able to solve on average 50 floors.

\begin{figure}[h]
    \centering
    \includegraphics[width=1.0\textwidth]{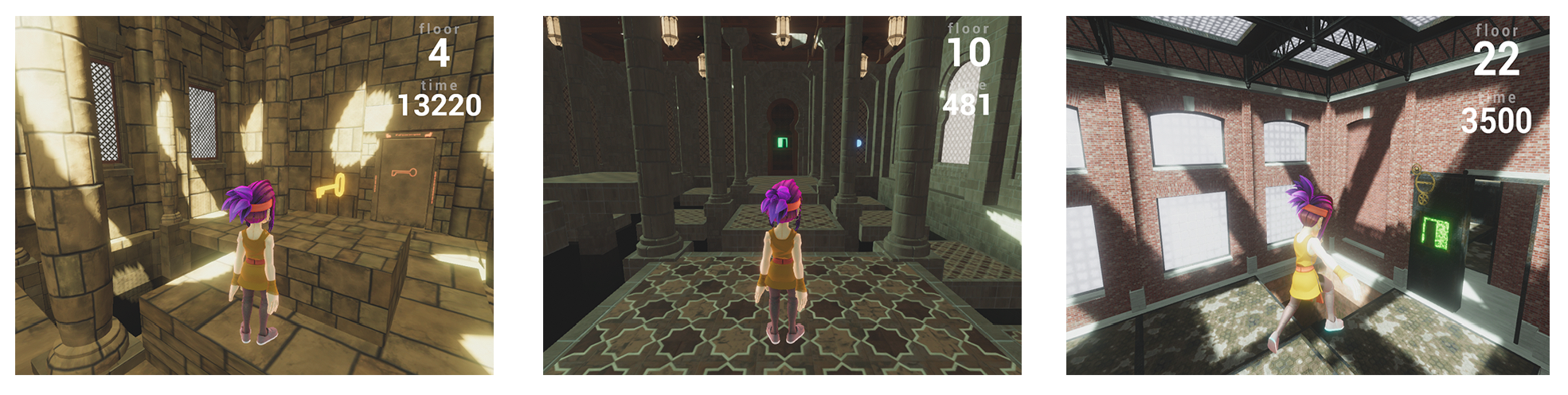}
    \caption{Examples of three floors generated in the Obstacle Tower environment.}
    \label{fig:ote}
\end{figure}

Concurrent with the publication of the baseline results reported in the original work, an open competition was held where research teams competed to train agents which could solve Obstacle Tower \footnote{https://www.aicrowd.com/challenges/unity-obstacle-tower-challenge}. These agents were evaluated on five held-out instances of Obstacle Tower not available during training. After six months of open contest, the top entry was able to solve an average of nineteen floors on the five held-out towers. This corresponds to better than naive human-level performance, but still well below expert human play, or optimal performance. 

In a blog post~\cite{nicholOT}, the top-scoring participant outlines their approach which consists of a creative combination of various RL and imitation learning modules as well as cleverly constructed human demonstrations and state augmentations; an invocation of the complexity of Obstacle Tower. This serves as an example of the role novel environments can serve in promoting the development of novel algorithms. Table~\ref{table:otc} contains results from the top six competitors. We encourage researchers who evaluate their algorithms on Obstacle Tower to compare to the results below in addition to those of the original work \cite{juliani2019obstacle}. 

\begin{table}[h!]
\centering
\begin{tabular}{|c c c c|} 
 \hline
 \textbf{Place} & \textbf{Contestant Name} & \textbf{Average Floors} & \textbf{Average Reward} \\ [0.5ex] 
 \hline
 1st & Alex Nichol & 19.4 & 35.86 \\
 2nd & Compscience.org & 16 & 28.7 \\
 3rd & Songbin Choi & 13.2 & 23.2 \\
 4th & Joe Booth & 10.8 & 18.06 \\
 5th & Doug Meng & 10 & 16.5 \\
 6th & UEFDL & 10 & 16.42 \\
 \hline
\end{tabular}
\caption{Performance on Obstacle Tower test-phase of top six entries in Obstacle Tower Challenge}
\label{table:otc}
\end{table}

\section{Potential for Future AI Research}\label{future_research}
As alluded to in the previous section, we believe there are a number of extremely valuable research directions that are hindered by the current standard benchmark problems. Working in these directions necessarily incurs additional overhead by forcing the researcher to create their own suitable environments~\cite{pathak2019morph} which can be a substantial burden if the tools of a general platform are unavailable. In this section, we highlight how the use of the Unity game engine can expedite research progress in lagging areas critical to the fields of AGI and human-AI interaction.

\subsection{Effective Learning Environments}
It has been argued in recent work that generating effective and diverse learning environments (often as a co-evolutionary process involving agent and environment) is a critical component for developing artificial general intelligence~\cite{wang2019poet,clune2019aiga}. Furthermore, other lines of research argue that procedural generation of environments and measuring success of an algorithm using a train/test split is a principled way of understanding the generalization and robustness of learning algorithms~~\cite{cobbe2019quantifying,justesen2018illuminating}.

As discussed in Section~\ref{UnityPlatform}, Unity environments are highly programmable via a straightforward C\texttt{\#} scripting system. This enables a simple way to control changing environment dynamics and dynamically create and destroy new entities (i.e. GameObjects), two critical components of an evolving environment. Furthermore, it is very natural for Unity environments to be parameterized and procedurally generated. This flexibility is uncommon among the platforms currently in use today. Additionally, Unity also has a large and active development community so that creating new and diverse environments is easy with an expansive array of off-the-shelf assets.

\subsection{Human-in-the-loop Training}
Leveraging human input to guide the learning process is desirable as exploiting a human's domain expertise speeds up learning and helps the agent learn to behave in a manner aligned with human expectations.  A number of training frameworks have been studied in the literature~\cite{zhang2019guidance} such as learning to imitate expert trajectories~\cite{ho2016generative}, humans providing evaluative feedback to the agent~\cite{knox2008tamer}, or humans manipulating the agent's observed states and actions~\cite{abel2016HIL}. The success of the latter two families of algorithms is in large part dependent on how the human interfaces with the agent {\it during learning} which is very difficult or impossible with the current set of platforms. On the other hand, imitation learning is a significantly more mainstream field of research which we hypothesize is partly because recording expert demonstrations requires very little extra functionality from the platforms themselves. An alternate line of work investigates how to design agents that don't learn to avoid being interrupted by humans given that it may prevent them from receiving future reward~\cite{orseau2016safely}.

Training within a visual environment editor, such as the Unity Editor, allows for an interactive and collaborative learning process between the human and agent. The editor offers real-time access to the training scene so that a human can interact with the agent during training simply via mouse clicks. Possible interventions include but are not limited to pausing the scene, dragging GameObjects within the scene, adding or removing GameObjects to the scene, and even assuming control of the agent through keyboard commands. This functionality will make the actual act of administering feedback and modifying the environment during training straightforward lifting a major burden in this field of research.

\subsection{Training Agents Alongside Humans}
Developing games with the assistance of artificial agents has a long history in the domain of game design~\cite{zhao2019wining}. Of particular value to the game development community is the ability to train flexible behaviors for non-playable characters (NPC) as either friend or foe to the player. Contained within this training dynamic is the under-explored research problem of training agents to be challenging to humans but not so dominant that the human does not engage in future contest. This may not align with an RL agent's goal of learning an optimal strategy. Training agents to perform at a particular player strength has been achieved via behavioral cloning and conditioning the policy on an estimate of the skill of the player that generated the demonstration~\cite{vinyalsAlphastar}. Thus, when a particular strength is desired, the network can be conditioned. However, we believe there to be novel RL formulations which seek to optimize the standard expected return within an episode {\it but also must optimize the number of expected future episodes}. A formulation of this sort could lead to a new family of RL algorithms and have implications for existential concerns for AI such as the value alignment problem~\cite{bostrom}.  

It is not trivial to investigate the training scenario where agents play against (or in cooperation with) humans robustly or at scale.  However, Unity's WebGL build option enables users to deploy Unity games to a browser. Thus, agent-human interaction can be studied at scale as humans play with or against an agent in a web browser game. As a side note, training agents against many humans with different play styles will also improve generalization and robustness of the learned policy~\cite{cobbe2019quantifying}.

\section{Conclusion and Future Directions}
In this paper, we introduce the notion of a {\it general platform} for environment creation and analyze the capabilities of the Unity engine with the Unity ML-Agents Toolkit as a member of this class.  To that end, we discussed the desirable complexity and computational properties of a simulator for the continued development of AI and used that criteria to propose a novel taxonomy of existing simulators and platforms.  From this analysis, we argued that the current set of platforms are insufficient for long-term progress and proposed modern game engines as the natural next step. We then discussed the Unity game engine and Unity ML-Agents Toolkit in this context and highlighted the key role it has already played in spurring innovation within the field.  Finally, we surveyed a subset of the research that using an engine like Unity enables but is currently burdensome to pursue due to the inflexibility of the current platforms.

There exist numerous other directions for future research in addition to those discussed in Section~\ref{future_research}. In addition to researchers, the Unity ML-Agents Toolkit is also intended to be used by game developers who are not necessarily machine learning experts. The notoriously tedious process of tuning hyperparameters may be insurmountable in some cases for a non-expert. We plan to introduce intuitive UI abstractions for navigating the iterative process of tuning an algorithm such as methods to tweak reward functions, defining observations, and defining actions as well as other aspects of algorithm design. Finally, other future work includes improving the Unity engine and the Unity ML-Agents Toolkit in both performance and breadth.

\section{Acknowledgements}

We would like to thank Jeff Shih, Anupam Bhatnagar, Adam Crespi, Deric Pang, Sachin Dharashivkar, Ruoping Dong, Sankalp Paltro, and Esh Vckay for their contributions to the Unity ML-Agents Toolkit; Praneet Dutta, Christopher Lu, and Cesar Romero for their feedback during the initial toolkit design process; Trevor Santarra, Vladimir Oster, Samuel Warren, YouCyuan Jhang, Joe Ward, Catherine Morrison, and Jose De Oliveira for their feedback on a draft of this paper.

\vskip 0.2in
\bibliographystyle{apalike}
\bibliography{bibliography}

\end{document}